\documentclass[conference,final]{IEEEtran}
\usepackage{graphicx}  % for pdf, bitmapped graphics files
\usepackage{epsfig} % for postscript graphics files
\usepackage[cmex10]{amsmath} % assumes amsmath package installed
\usepackage{amsfonts}
\usepackage{amsmath}
\usepackage{amssymb}
\usepackage{amsthm} %command for \proof
\usepackage{bm}     %use \bm instead of \mathbf
\usepackage{algorithm}
\usepackage[noend]{algpseudocode}
\usepackage[]{units}
\usepackage{placeins}
\usepackage{subcaption}
\usepackage{epstopdf}

\usepackage{tikz}

\newcommand{\citep}[1]{\cite{#1}}

\graphicspath{{../../Figures/}{.}}
\begin{document}

%
% paper title
% can use linebreaks \\ within to get better formatting as desired
%\title{Dynamic Resource Allocation in Virtualized Computing Pipelines}
\title{Automated Identification of Drug-Drug Interactions in Pediatric Congestive Heart Failure Patients}

\author{\IEEEauthorblockN{Daniel Miller}
\IEEEauthorblockA{Final Report\\
CS 221, Stanford University\\
Email: danielrm@stanford.edu, SUNet ID: 005579372}
}

% make the title area
\maketitle

%%%%%%%%%%%%%%%%%%%%%%%%%%%%%%%%%%%%%%%%%%%%%%%%%%

%%%%%%%%%%%%%%%%%%%%%%%%%%%%%%%%%%%%%%%%%%%%%%%%%%
\section{INTRODUCTION} % and initial literature review
%%%%%%%%%%%%%%%%%%%%%%%%%%%%%%%%%%%%%%%%%%%%%%%%%%

Congestive Heart Failure, or CHF, is a serious medical condition that can result in fluid buildup in the body as a result of a weak heart.  When the heart can't pump enough blood to efficiently deliver nutrients and oxygen to the body, kidney function may be impaired, resulting in fluid retention.  CHF patients require a broad drug regimen to maintain the delicate system balance, particularly between their heart and kidneys.  These drugs include ACE inhibitors and Beta Blockers to control blood pressure, anticoagulants to prevent blood clots, and diuretics to reduce fluid overload.  Many of these drugs may interact, and potential effects of these interactions must be weighed against their benefits.

For this project, we consider a set of 44 drugs identified as specifically relevant for treating CHF by pediatric cardiologists at Lucile Packard Children's Hospital.  This list was generated as part of our current work at the LPCH Heart Center.

The goal of this project is to identify and evaluate potentially harmful drug-drug interactions (DDIs) within pediatric patients with Congestive Heart Failure.  This identification will be done autonomously, so that it may continuously update by evaluating newly published literature.

%%%%%%%%%%%%%%%%%%%%%%%%%%%%%%%%%%%%%%%%%%%%%%%%%%
\section{GOALS AND DATA}
%%%%%%%%%%%%%%%%%%%%%%%%%%%%%%%%%%%%%%%%%%%%%%%%%%

\subsection{Automated Literature Analysis}

In the first stage, I will use natural language processing to extract drug interactions from medical research paper abstracts.  The input to this stage will be raw abstract texts related to a DDI, like the one shown in Figure \ref{fig:ex_abstract}.

\begin{figure}[tb]
\fbox{\begin{minipage}{\linewidth}
Bumetanide and furosemide in heart failure. We assessed the handling of and response to oral bumetanide (1.0 and 2.0 mg) and to furosemide (40 and 80 mg) in 20 patients with stable, compensated congestive heart failure (CHF), comparing the two drugs and, in addition, examining differences from normal subjects. Bumetanide and furosemide were similar in time course of absorption, but patients with CHF had considerably prolonged absorption compared to normal subjects causing attainment of lower peak concentrations of drug. In both CHF and normal subjects, more bumetanide than furosemide was absorbed. The elimination half-life of furosemide was approximately twice that of bumetanide, and both were about two times longer than respective values in normal subjects. “Dose”-response curves were shifted downward from normal with both drugs. In patients with CHF, overall response did not differ between bumetanide and furosemide. The two drugs exhibit subtle differences, the clinical importance of which appears to be negligible from this study. Importantly, however, both drugs showed delayed absorption causing attainment of peak urinary excretion rates of diuretic two- to threefold lower than in normal subjects. This effect along with the abnormal responsivity of the tubule may contribute to the “resistance” to oral doses of diuretics observed clinically even though no quantitative malabsorption of drug occurs.
\end{minipage}}
\caption{An example drug-drug interaction abstract \cite{brater1984bumetanide}.  Furosemide and Bumetanide are both diuretics, used to dehydrate the patient.}\label{fig:ex_abstract}
\end{figure}

As an initial evaluation metric (labels), I have used DrugBank, an established resource for drug data, which includes drug-drug interaction and much more.  DrugBank was assembled over several years by a team of archivists and annotators that included two accredited pharmacists, a physician and three bioinformaticians with dual training in computing science and molecular biology/chemistry \cite{wishart2006drugbank}.  One goal of my project is to design a automated system capable of replicating, and extending the DDI component of this work.

All abstracts used in this project come from the PubMed Central Open Access Subset \cite{PubMed}.  These data are all openly available in .xml format via FTP, and currently (as of November 2016) include a total of 663,597 papers.  These abstracts required significant filtering to narrow down to only those papers concerned with drug-drug interactions, specifically for drugs relevant to the cardiac patient population.  To perform this filtering, I fist considered the set of 44 cardiac drugs.  Using the DrugBank database, I identified a superset of 1781 drugs known to have interactions with at least one of the cardiac drugs.  Any abstract that contained mention of at least one of these 1781 drugs is defined as a ``Cardiac Abstract'' and included in the analysis.

This filtering process left 69,713 remaining abstracts, roughly 10\% of the initial total.  I used the PubMed Parser tookit to extract the abstracts from the xml formatted dataset \cite{PubMedParser}.

\subsection{Patient Analysis}

The second goal of my project is to identify potential drug interactions for a sample Congestive Heart Failure patient population, based on the output from the previous stage.  For this project, I will be using an anonymized extract of the full Medical Administration Record (MAR) for ~60 current and former CHF patients at the Lucile Packard Children's Hospital (LPCH).  The specific inputs to this stage are:

\begin{enumerate}
	\item The drug interactions generated by the previous stage
	\item The Medical Administration Record (MAR) for the CHF patients (drug name and admin time).
\end{enumerate}

\begin{figure}[tb]
\includegraphics[width=\linewidth]{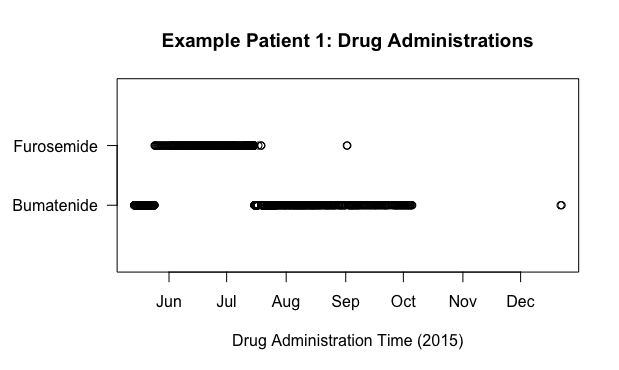}
\caption{An example Medication Administration Record (MAR).}\label{fig:ex_mar}
\end{figure}

An example of a single patient's Furosemide and Bumetanide administration records is shown in Figure \ref{fig:ex_mar}.  As we can see from the plot, these drug administrations overlap, creating time windows where the patient might experience detrimental DDI effects.

The output of this stage is a set of time-ranges where we predict potential DDIs.  For example, from the record shown in Figure \ref{fig:ex_mar}, we see that Patient 1 received Furosemide and Bumatenide in July 2015, with potential effect of decrease kidney function, and generate the following coded output:

\begin{itemize}
	\item ((Furosemide, Bumatenide), ("2015-07-15",\\ "2015-07-19"), "Reduced Kidney Function")
\end{itemize}

The evaluation metric for this stage is not yet clearly defined.  Once the system has generated specific output hypotheses like the one shown above, I will request specific clinician feedback from the LPCH cardiologists.  Ideally, we would be able to tie each effect to a known measurable symptom.  Continuing the example, reduced kidney function is often recognized by an increase in creatinine in the kidneys.  Creatinine labs are taken frequently for the CHF patients.

This section of the project is beyond the scope of this class, and has been deferred pending improvements in the algorithm performance for the automated literature analysis.  Future work will include implementation of DDI identification for current LPCH patients, and automated warning messages to physicians when a clearly defined risk is achieved.  One significant issue to be tackled is properly balancing the false negative vs false positive rate to suppress false alarms, and maintain sensitivity.

%%%%%%%%%%%%%%%%%%%%%%%%%%%%%%%%%%%%%%%%%%%%%%%%%%
\section{Model}
%%%%%%%%%%%%%%%%%%%%%%%%%%%%%%%%%%%%%%%%%%%%%%%%%%

For the time being, we consider only part $A$, the automated literature analysis.  The general use case of the proposed model may be summarized as follows:

\begin{enumerate}
	\item The user submits a query of the following form:  ``Do the drugs $A$ and $B$ have any potentially harmful interactions, and if so, how may they interact?''
	\item A large set of medical publication abstracts are provided, with each related to either drug $A$, $B$, or both.
	\item Our model runs on these inputs, and returns the drug-drug interactions, if any exist.
\end{enumerate}

To generate a model fulfilling this form, we take the following steps.

\subsection{Labeling}
To begin, we generate all pairwise drug combinations.  We begin by considering the 44 pediatric cardiology drugs $\mathcal{D}_{cardiac}$.  For each cardiac drug $d_c \in \mathcal{D}_{cardiac}$, we extract all pairwise drug interactions $(d_c, d)$ from the DrugBank database $\mathcal{D}_{drugbank}$.  The union of all considered drugs $d$ forms the set of all drugs $\mathcal{D}_{all}$ considered in this analysis.

For each cardiac drug $d_c \in \mathcal{D}_{cardiac}$, we create a sample for every other drug in $d_o \in \mathcal{D}_{all}$.  We assign a indicator label if two drugs interact as follows:

\[
\mbox{label}\left( (d_c, d_o) \right) = \left\{
\begin{array}{ll}
1 & \mbox{  if } (d_c, d_o) \ \in \mathcal{D}_{drugbank}\\
0 & \mbox{otherwise} \\
\end{array}
\right.
\]

This label is used for the binary classification supervised learning problem, in which we predict whether two drugs interact, but not the nature of their interaction.  We also consider the type of interaction, which is also provided by the DrugBank database.  By replacing the drug names in each interaction, we obtain 53 interaction types $i\in \mathcal{I}_{drugbank}$ of the following form:

\textit{Ex: "The serum concentration of (~drug~) can be increased when it is combined with (~drug~)."}

These interaction types will be used to label the supervised multiclass classification problem in which we will predict the actual type of interaction between two drugs, if any exists.

\begin{table}[]
\centering
\caption{Data Counts}
\label{table:data_counts}
\begin{tabular}{lllll}
\textbf{Cardiac Abstracts} & 69713 \\
\textbf{Cardiac Drugs}     & 44    \\
\textbf{Cardiac Drugs in Abstracts}     & 36    \\
\textbf{Related Drugs}     & 1781  \\
\textbf{Drug Interactions} & 63450 \\
\textbf{Types of Drug Interactions} & 53 \\
\textbf{Drug Interactions between 2 cardiac drugs} & 218 \\
\textbf{Average Drugs per Abstract} &  1.3 \\
\textbf{Maximum Drugs per Abstract} &  80 \\
\textbf{Average Words per Abstract} & 149.5  \\
\textbf{Average Count per Words per Abstract} & 1.9  \\
\textbf{Total Distinct Words in all Abstracts} & 53338  \\
\end{tabular}
\end{table}

\subsection{Test/Train/Validation Split}

For this particular problem, a particularly careful definition of the validation splits is worth examination.  Because each abstract may consider one or more drugs (see Table~\ref{table:data_counts}), we would like to use the information each abstract for more than one interaction (sample).  Therefore, our samples will not be independent, so a validation set randomly sampled from the interactions could be highly correlated with the training set.  We therefore take precautions to eliminate this dependence by splitting on interactions via the following procedure:

\begin{algorithm}
\caption{Train/Test/Validation Split Algorithm}
\begin{algorithmic}[1]
	\State Split all abstracts into training (80\%) and testing (20\%) sets.	
	\State Split all interactions into training (80\%) and testing (20\%) sets.
	\State Split training set into training (80\%) and validation/dev (20\%) sets.
	\For {split in (train, test, validation)}	
		\For {each pairwise interaction $(d_1, d_2)$	in split}
			\For {each abstract in split}
				\If {$d_1$ or $d_2$ in abstract}
					\State Assign abstract to the interaction $(d_1, d_2)$
				\EndIf
			\EndFor
		\EndFor
	\EndFor
\end{algorithmic}
\end{algorithm}  

In this way, DDI's in the training set may share abstracts, but not with any DDI's in the test or validation set.

\subsection{Feature Extraction}

We next extract features for each DDI from the abstracts.  So far, we have considered two methods for performing feature extraction.

\subsubsection{Word Counts} 
For the moment, we are simply using a sparse vector word counter for each abstract assigned to an interactions, and summing over all abstract counters to get a single feature vector for each interaction.  Refer to Table~\ref{table:data_counts} for summary information on the counts of words in the abstracts.

The current model limitation is rooted in the scaling.  Many abstracts are hundreds of words long, and make mention of dozens of drugs.  Our final sparse word counter considers 53338 words across 63450 interaction samples.  To allow some methods to run, it was necessary to drop all but the 100 most common words to allow for a reasonable runtime.

\subsubsection{GloVe}

To address the scaling issue, I have also made use of GloVe: Global Vectors for Word Representation~\cite{pennington2014glove}.  This implementation is in the very early stages, and thus far has not demonstrated notable accuracy.  In the interest of time, we used the Wikipedia 2014 + Gigaword 5 pre-trained word vectorizations.  This contained vectorizations for 31774 of the words across all cardiac abstracts, but did not contain pre-trained vectors for 21560 of the words, most of which were medically-specific.  In the future, I will re-build a medically-specific GloVe vectorization from the entire PubMed OA subset, including both abstract and paper body test.

To move from word to sentence and full abstract vectors, I simply used a sum over the words in the abstracts, dropping all stop words, and weighted by term frequency.  This is clearly non-ideal, as it does not take into account word order or semantics/context, but was done in the interest of time.  Even this simple method took nearly a day of computation time on a personal machine.  Fortunately, using pre-trained word vectors to analyze each abstract is a fully parallelizable procedure, which will be taken advantage of going forward.  I am currently researching possible algorithms to handle a highly variable collection of abstracts, each of variable length.  One option is to pre-process using Fixed-size Ordinally-Forgetting Encoding (FOFE)~\cite{zhang2015fixed}.

In the future, we will consider more sophisticated features, including pairwise indicators, n-grams, and adjacent pairs.

\subsection{Supervised Learning: Evaluation and Model Selection}

\begin{table}[]
\centering
\caption{Preliminary Model Results}
\label{table:preliminary_results}
\begin{tabular}{lllll}
Method & Sens. & Spec. & PPV & NPV \\ \hline
\textbf{Raw linear SVM (dev)} & 1 & 0 & 0.779 & N/A \\
\textbf{Undersampled SVM (dev)} & 0.994 & 0.017 & 0.968 & 0.093 \\
\textbf{Raw CNN (train)} & 1 & 0 & 0.782 & 1 (err)    \\
\textbf{Undersampled  CNN (train)} & 0.733 & 0.536 & 0.850 & 0.359 \\
\end{tabular}
\end{table}

Thus far, I have implemented a variety of distinct supervised learning algorithms.  Each is trained exclusively on the training split, and all model tuning is performed by splitting the training set into further validation sets and cross-validation folds.  Each model's performace is evaluated and compared on the validation set.  As more methods are considered, and further tuning is performed, this pattern will continue.  The final model selection will be based on performance on the validation/dev set.  This final model will be re-tuned on the combined train/validation sets, and only then will the final performance metrics be evaluated on the test set.

The first model considered was a linear SVM using the word count features, due to its convenient scalability, and allowance for cheap feature extensions via kernel methods.  I implemented this model using the \emph{e1071} package in R, since it supports sparse predictor matrices.  This allowed me to consider all 53338 word counts.  Initial results followed the typical result for SVM classifiers on high-dimensional imbalanced-label datasets.  Somewhat unsurprisingly, the SVM model classified all samples to the most common class (0 -- no interaction), as this yielded a relatively high accuracy.

To counteract the imbalancing, I tried undersampling the training set to a balanced number of each label.  This led to minor improvements as seen in Table~\ref{table:preliminary_results}.  I expect further improvements to come from tuning and kernel selection, but I am not convinced that SVM is an appropriate model for this problem.

\begin{figure}[tb]
\includegraphics[width=\linewidth]{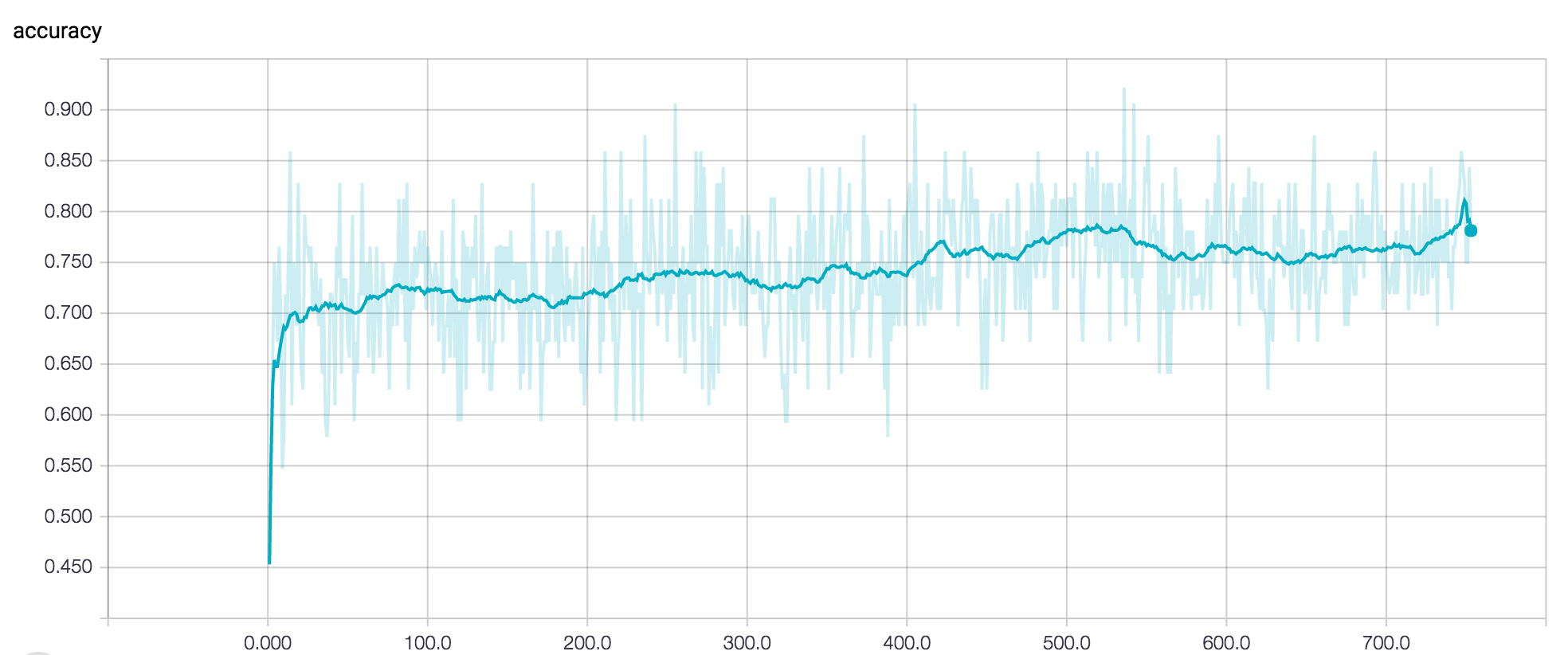}
\caption{CNN training accuracy per iteration.}\label{fig:cnn_accuracy}
\end{figure}

The second model considered was a convolutional neural network implemented in TensorFlow with a re-purposed fork of the \emph{Convolutional Neural Networks for Sentence Classification} toolkit written and published by Yoon Kim \cite{DBLP:journals/corr/Kim14f}.  As CNNs do not scale as nicely as SVMs, it was necessary to reduce the extremely large feature set.  As a preliminary selection, I just selected out the counts of the 100 most common words.  This model also initially experienced the same issues that SVM had with the imbalanced labels.  However, using the same undersampling trick yielded much better results, leading to the promising, if not immediate useful metrics seen in Table~\ref{table:preliminary_results}.  One advantage of this model is the ability to continuously tune over time.  Figure~\ref{fig:cnn_accuracy} shows how the training accuracy of the CNN increases as we continue to iterate the model.

\begin{figure}[tb]
\includegraphics[width=\linewidth]{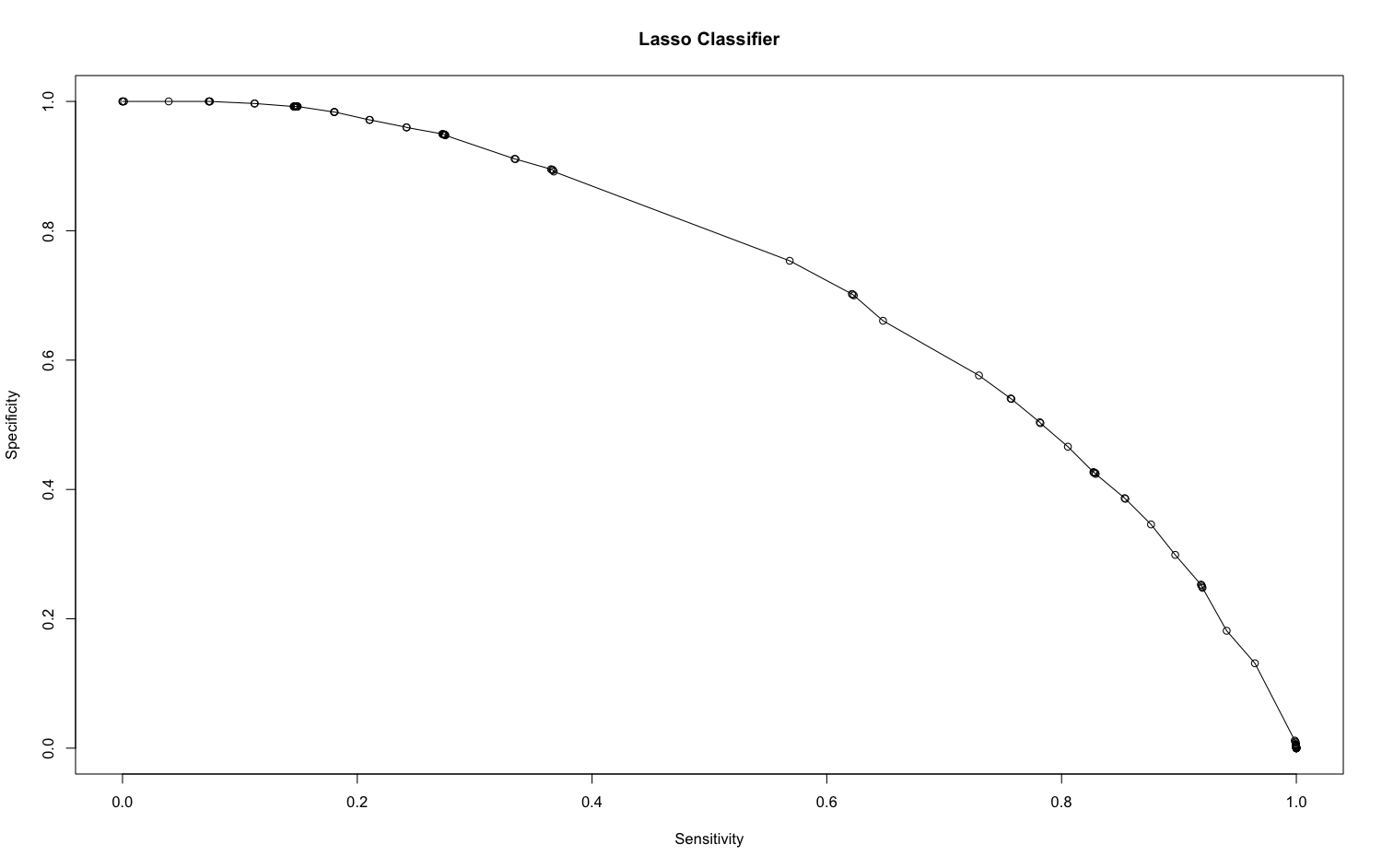}
\caption{Lasso Classifier with word counts validation/dev sensitivity-specificity curve}\label{fig:lasso_wordcount_all}
\end{figure}

I next tried to directly address the feature set size and enforced sparsity by introducing L1 regularization via the Lasso method.  Thus far, this method has been the most successful by a significant margin.  The full sensitivity-specificity curve is shown in Figure~\ref{fig:lasso_wordcount_all}.  The regularization parameter was tuned via cross-validation with the reward function being area-under-the-curve (AUC).

\begin{figure}[tb]
\includegraphics[width=\linewidth]{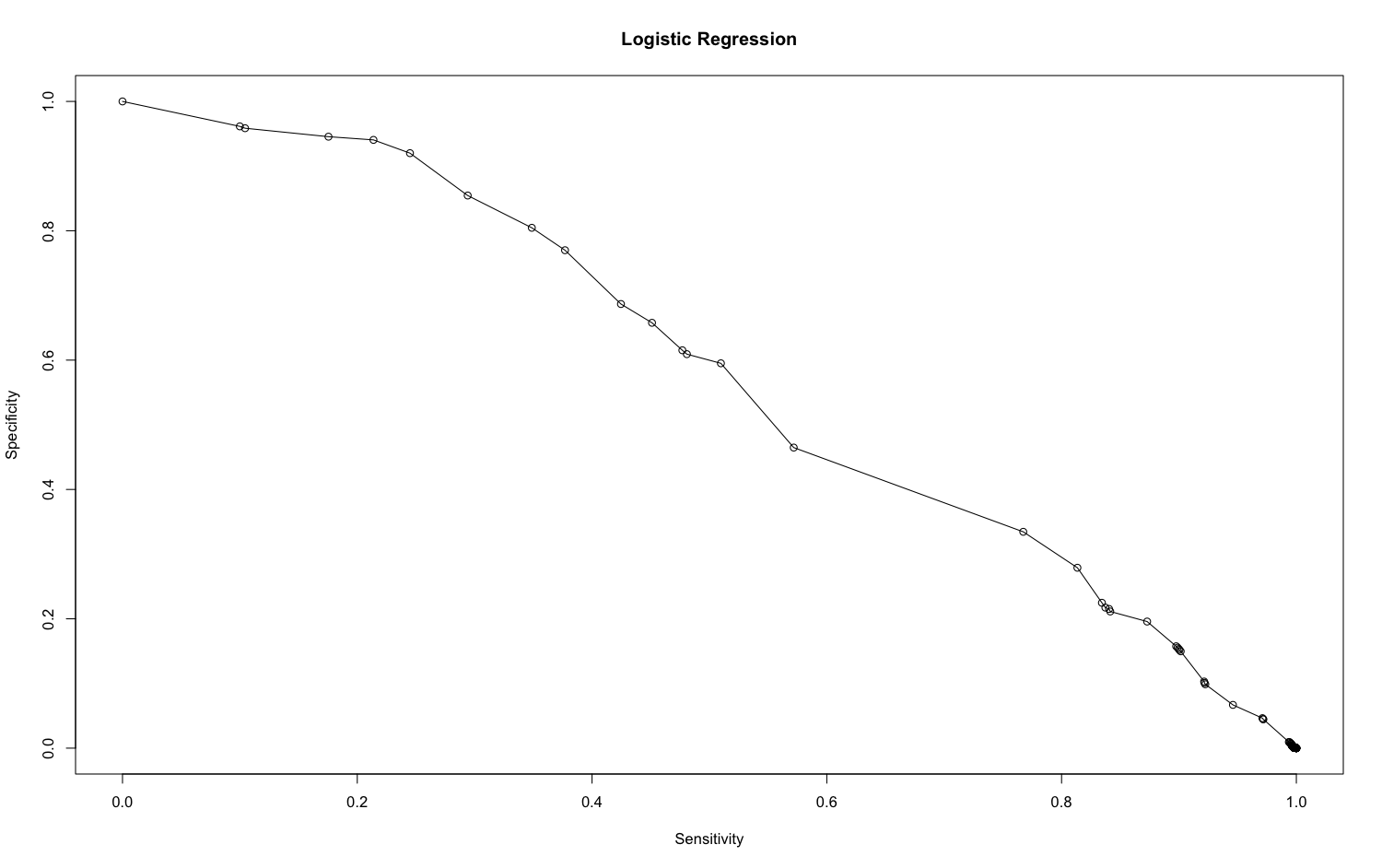}
\caption{Logistic Regression with GloVe validation/dev sensitivity-specificity curve}\label{fig:logistic_glove}
\end{figure}

\begin{figure}[tb]
\includegraphics[width=\linewidth]{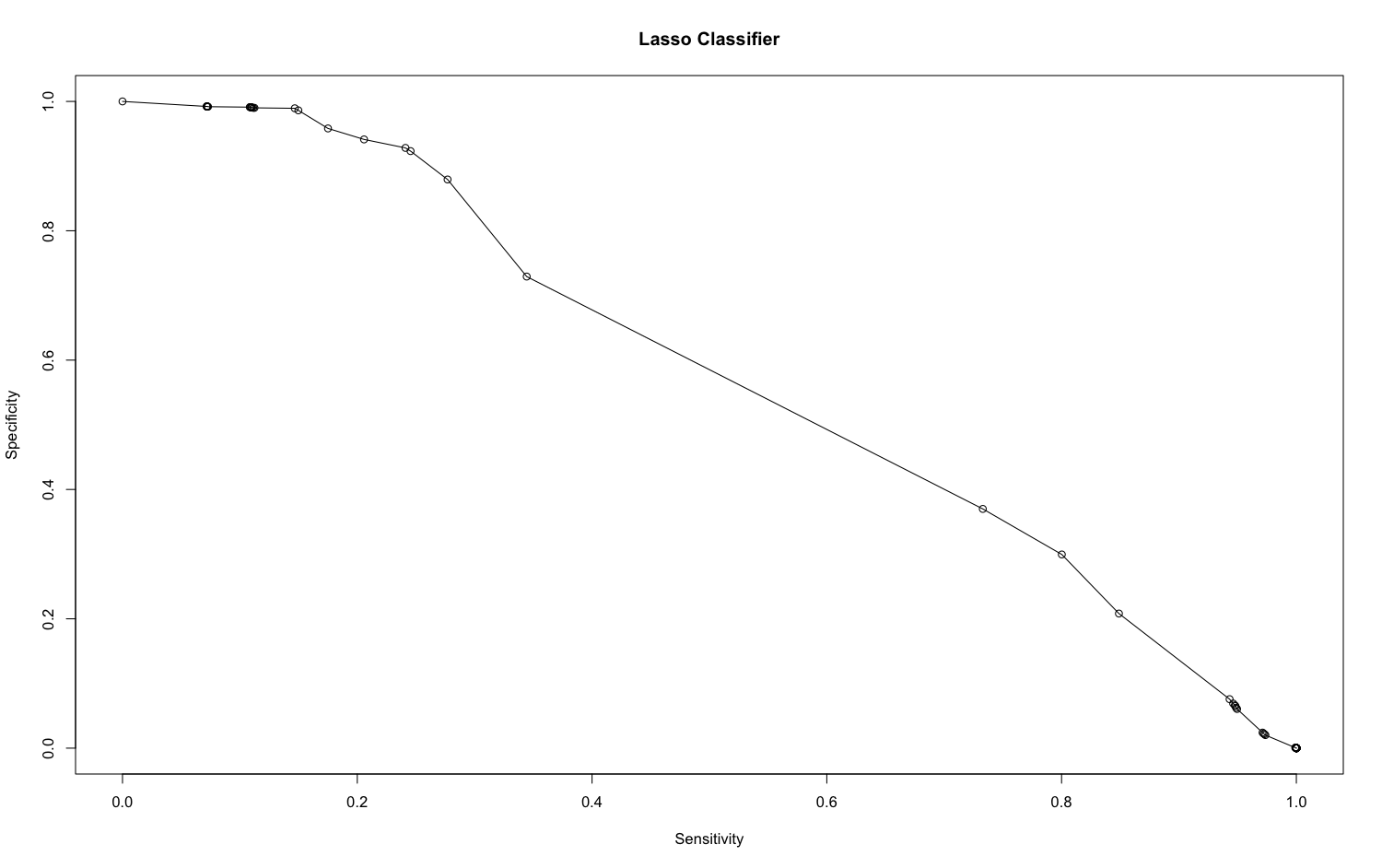}
\caption{Lasso Classifier with GloVe validation/dev sensitivity-specificity curve}\label{fig:lasso_glove}
\end{figure}

The final methods both used the GloVe vectorizations rather than the word counts.  While performing better than random, neither method approached the results of the wordcount Lasso implementation.  The first method, shown in Figure~\ref{fig:logistic_glove} is a simple logistic regression applied to the GloVe vectors. The second method, shown in Figure~\ref{fig:lasso_glove} is a Lasso classifier applied to the GloVe vectors, again tuned via cross-validation for AUC.

%%%%%%%%%%%%%%%%%%%%%%%%%%%%%%%%%%%%%%%%%%%%%%%%%%
\section{Future Work and Conclusions}
%%%%%%%%%%%%%%%%%%%%%%%%%%%%%%%%%%%%%%%%%%%%%%%%%%

The preliminary methods and results presented in this paper demonstrate the viability of the goals of this project.  In particular, the Lasso method on word counts points towards using feature selection in conjunction with a broad, automatically generated feature space.  The primary limitation of the project thus far has been scalability.  Fortunately, these  issues are focused around highly parallelizable algorithms, and may be addressed going forward by GPU implementation.  In the same vein, more efficient implementations will allow for more sophisticated NLP algorithms to perform feature extraction from the set of abstracts associated with each drug-drug interaction.

In summary, our results indicate that there is a useful information about drug-drug interactions for congestive heart failure patients that may be extracted autonomously from the medical literature, with no external clinical knowledge.

%%%%%%%%%%%%%%%%%%%%%%%%%%%%%%%%%%%%%%%%%%%%%%%%%%
%\section*{APPENDIX}
%%%%%%%%%%%%%%%%%%%%%%%%%%%%%%%%%%%%%%%%%%%%%%%%%%

%%%%%%%%%%%%%%%%%%%%%%%%%%%%%%%%%%%%%%%%%%%%%%%%%%%%%%%%%%%%%%%%
%\addtolength{\textheight}{-2.5cm}   % This command serves to balance the column lengths
%                                  % on the last page of the document manually. It shortens
%                                  % the textheight of the last page by a suitable amount.
%                                  % This command does not take effect until the next page
%                                  % so it should come on the page before the last. Make
%                                  % sure that you do not shorten the textheight too much.
%%%%%%%%%%%%%%%%%%%%%%%%%%%%%%%%%%%%%%%%%%%%%%%%%%%%%%%%%%%%%%%%

%%%%%%%%%%%%%%%%%%%%%%%%%%%%%%%%%%%%%%%%%%%%%%%%%%
% References Section
%%%%%%%%%%%%%%%%%%%%%%%%%%%%%%%%%%%%%%%%%%%%%%%%%%
\pagebreak
\bibliography{CS221_References}

% Generated by IEEEtran.bst, version: 1.13 (2008/09/30)
\begin{thebibliography}{1}
\providecommand{\url}[1]{#1}
\csname url@samestyle\endcsname
\providecommand{\newblock}{\relax}
\providecommand{\bibinfo}[2]{#2}
\providecommand{\BIBentrySTDinterwordspacing}{\spaceskip=0pt\relax}
\providecommand{\BIBentryALTinterwordstretchfactor}{4}
\providecommand{\BIBentryALTinterwordspacing}{\spaceskip=\fontdimen2\font plus
\BIBentryALTinterwordstretchfactor\fontdimen3\font minus
  \fontdimen4\font\relax}
\providecommand{\BIBforeignlanguage}[2]{{%
\expandafter\ifx\csname l@#1\endcsname\relax
\typeout{** WARNING: IEEEtran.bst: No hyphenation pattern has been}%
\typeout{** loaded for the language `#1'. Using the pattern for}%
\typeout{** the default language instead.}%
\else
\language=\csname l@#1\endcsname
\fi
#2}}
\providecommand{\BIBdecl}{\relax}
\BIBdecl

\bibitem{brater1984bumetanide}
D.~C. Brater, B.~Day, A.~Burdette, and S.~Anderson, ``Bumetanide and furosemide
  in heart failure,'' \emph{Kidney international}, vol.~26, no.~2, pp.
  183--189, 1984.

\bibitem{wishart2006drugbank}
D.~S. Wishart, C.~Knox, A.~C. Guo, S.~Shrivastava, M.~Hassanali, P.~Stothard,
  Z.~Chang, and J.~Woolsey, ``Drugbank: a comprehensive resource for in silico
  drug discovery and exploration,'' \emph{Nucleic acids research}, vol.~34, no.
  suppl 1, pp. D668--D672, 2006.

\bibitem{PubMed}
``{PubMed Open Access Subset},''
  \url{https://www.ncbi.nlm.nih.gov\\/pmc/tools/openftlist/}, online; accessed:
  27-October-2016.

\bibitem{PubMedParser}
T.~Achakulvisut and D.~E. Acuna, ``{PubMed Parser},''
  \url{http://github.com/titipata/pubmed\_parser.}
  \url{http://doi.org/10.5281/zenodo.159504}, 2015, [Online; accessed
  16-November-2016].

\bibitem{pennington2014glove}
\BIBentryALTinterwordspacing
J.~Pennington, R.~Socher, and C.~D. Manning, ``Glove: Global vectors for word
  representation,'' in \emph{Empirical Methods in Natural Language Processing
  (EMNLP)}, 2014, pp. 1532--1543. [Online]. Available:
  \url{http://www.aclweb.org/anthology/D14-1162}
\BIBentrySTDinterwordspacing

\bibitem{zhang2015fixed}
S.~Zhang, H.~Jiang, M.~Xu, J.~Hou, and L.~Dai, ``The fixed-size
  ordinally-forgetting encoding method for neural network language models,''
  \emph{Proceedings of ACL}, 2015.

\bibitem{DBLP:journals/corr/Kim14f}
\BIBentryALTinterwordspacing
Y.~Kim, ``Convolutional neural networks for sentence classification,''
  \emph{CoRR}, vol. abs/1408.5882, 2014. [Online]. Available:
  \url{http://arxiv.org/abs/1408.5882}
\BIBentrySTDinterwordspacing

\end{thebibliography}
\bibliographystyle{IEEEtran}

%\begin{figure*}[h]
%\includegraphics[width=\linewidth]{CNN_graph_1.png}
%\caption{CNN graph from TensorFlow.}\label{fig:cnn_graph}
%\end{figure*}

% that's all folks
\end{document}